\documentclass[fullpaper,final]{nldl}

\paperID{40}

\vol{233}

\usepackage{mathtools}

\usepackage{graphicx}

\usepackage{booktabs}

\usepackage{enumitem}

\usepackage{algorithm}
\usepackage{algorithmic}

\usepackage[utf8]{inputenc}
\usepackage{url}
\usepackage{graphicx}
\usepackage{authblk}

\usepackage{algorithm}
\usepackage{algorithmic}

\usepackage{amsmath}
\usepackage{amsthm}
\usepackage{amsfonts}
\usepackage{adjustbox}
\usepackage{multirow}

\DeclareMathOperator*{\argmax}{arg\,max}

\usepackage{listings}
\usepackage{hyperref}
\usepackage{url}
\hypersetup{
  colorlinks,
  linkcolor = BrickRed,
  citecolor = NavyBlue,
  urlcolor  = Magenta!80!black,
}

\title{MDD-UNet: Domain Adaptation for Medical Image Segmentation with Theoretical Guarantees, a Proof of Concept}
\author[1,2]{Asbjørn Munk\thanks{Corresponding Author: asmu@di.ku.dk}}
\author[3]{Ao Ma}
\author[1,2]{Mads Nielsen}
\affil[1]{Pioneer Centre for AI}
\affil[2]{University of Copenhagen}
\affil[3]{Southwestern University of Finance and Economics}

\begin{document}
\maketitle

\begin{abstract}
The current state-of-the art techniques for image segmentation are often based on U-Net architectures, a U-shaped encoder-decoder networks with skip connections. Despite the powerful performance, the architecture often does not perform well when used on data which has different characteristics than the data it was trained on. Many techniques for improving performance in the presence of \textit{domain shift} have been developed, however typically only have loose connections to the theory of domain adaption. In this work, we propose an unsupervised domain adaptation framework for U-Nets with theoretical guarantees based on the Margin Disparity Discrepancy \cite{zhang2019} called the MDD-UNet. We evaluate the proposed technique on the task of hippocampus segmentation, and find that the MDD-UNet is able to learn features which are domain-invariant with no knowledge about the labels in the target domain. The MDD-UNet improves performance over the standard U-Net on 11 out of 12 combinations of datasets. This work serves as a proof of concept by demonstrating an improvement on the U-Net in it's standard form without modern enhancements, which opens up a new avenue of studying domain adaptation for models with very large hypothesis spaces from both methodological and practical perspectives. Code is available at \url{https://github.com/asbjrnmunk/mdd-unet}.
\end{abstract}

\section{Introduction}
In medical image analysis data distributions vary considerably across equipment, patient groups, and scanning protocols \cite{saat2022}.  Since labeling medical images typically involves labor-intensive participation of specialists, available labeled data is often limited. This is a key challenge in medical image segmentation \cite{orbes-arteaga2019}, since models typically fail at generalizing to data which is different from the specific setup of the training data, while manually labeling data from each new test domain is infeasible \cite{kamnitsas2017}.

\begin{figure}[tb]
\centering
\includegraphics[width=1\linewidth]{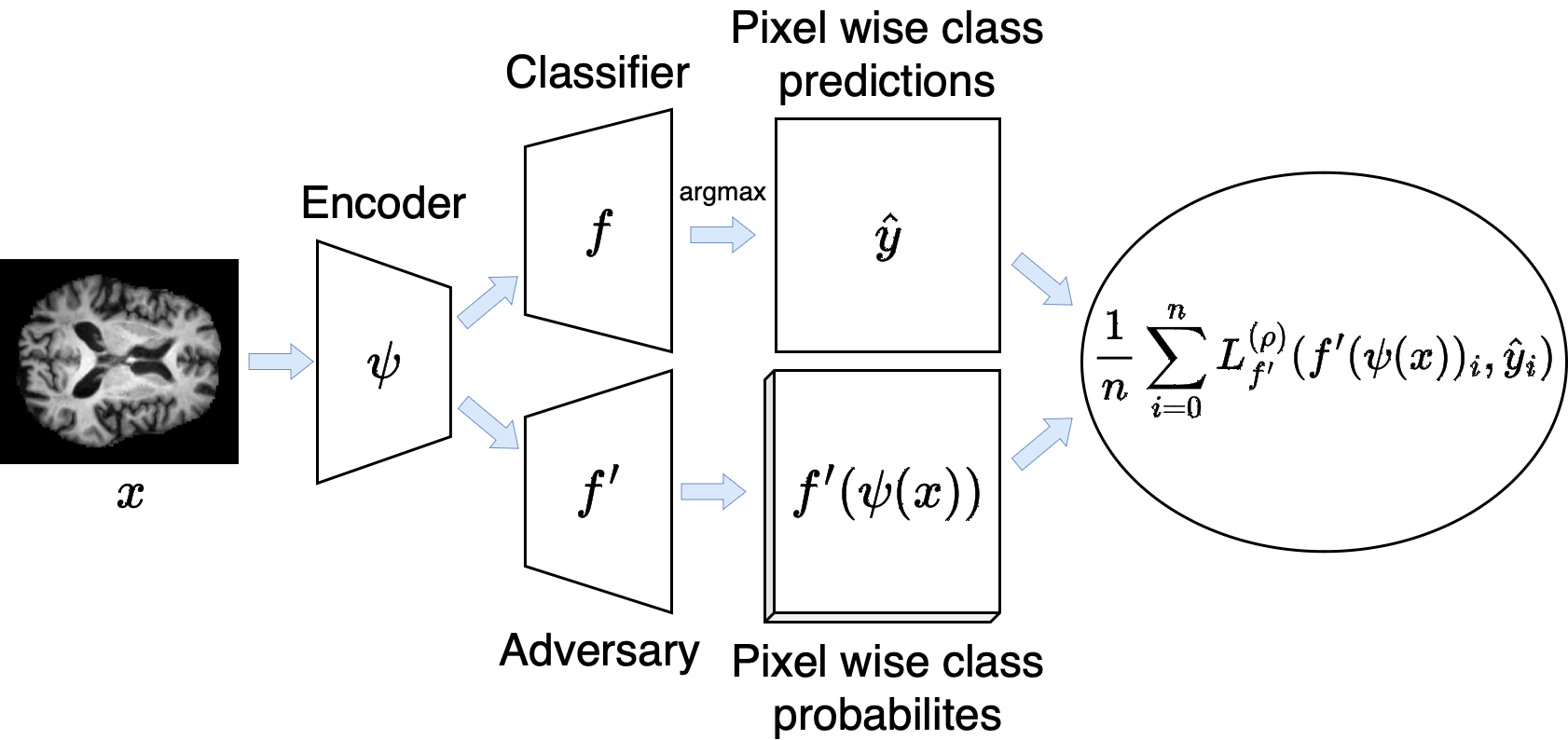}
\caption{\textbf{Margin Disparity}. The calculation of $\text{disp}_{D, \psi}^{(\rho)}(f', f)$, a measure of disparity between two classifiers, $f$ and $f'$. This measure importantly works for any classifier, enabling us to apply this directly to medical segmentation. The loss $L$ denotes the \textit{margin loss} up to some maximal margin $\rho > 0$. The final disparity is the average disparity over all pixels in the input.}
\label{fig:disparity}
\end{figure}

\begin{figure*}[tb]
\centering
\includegraphics[width=0.8\linewidth]{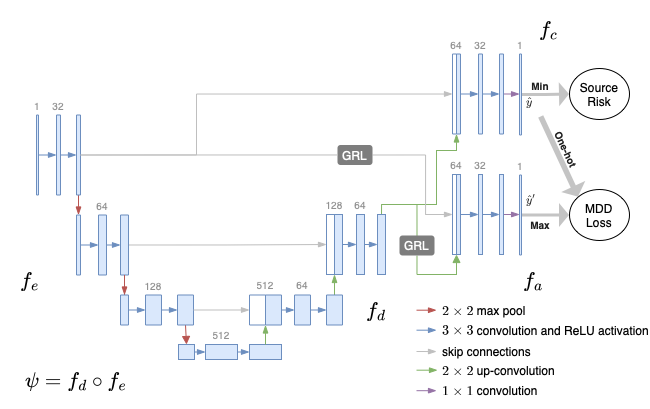}
\caption{\textbf{The proposed architecture}. The base model is a U-Net \cite{ronneberger2015} combined with the adversarial MDD architecture \cite{zhang2019}. Each box denotes an intermediate representation, with the amount of channels denoted by a small number above. GRL is a \textit{Gradient Reversal Layer} \cite{ganin}.}
\label{fig:architecture}
\end{figure*}

One solution to tackles this problem is unsupervised domain adaptation (UDA) \citep{guan2021}. In UDA the goal is to transfer knowledge learned from the source domain to a similar, yet distinct, target domain, while only assuming labels from the source domain.

While important technical advances have been developed in UDA, methodologies are generally not theoretically well-founded. At the same time, remarkable theoretical advances have been achieved in domain adaptation. In particular, the seminal work by Ben-David et al. \citep{bendavid2010} introduced the $\mathcal{H}\Delta\mathcal{H}$-divergence as a measurement of discrepancy between two distributions. This allowed rigorous learning bounds to be derived and showed, that to obtain good performance on the target domain, there is an intrinsic trade-off between performance on the source domain and the empirical $\mathcal{H}\Delta\mathcal{H}$-divergence.

Practical methods for domain adaptation seek to exploit this trade-off, for instance DANN \cite{ganin} employ an adversarial architecture, inspired by GAN \cite{goodfellow}, in which networks play a \textit{minimax game} seeking to learn a representation of the input where the source and target domains are indistinguishable, while performing well on the source domain. However, the theoretical foundation of DANN is limited to binary classifiers, meaning that for problems such as segmentation, the methodology lacks theoretical guarantees, since the hypothesis spaces of the max-player and min-player are distinctly dissimilar.

Zhang et al. \cite{zhang2019} remedy this by proposing a new distribution discrepancy measurement called Margin Disparity Discrepancy (MDD), allowing the derivation of generalization bounds comparable to those of Ben David et al. \cite{bendavid2010} based on scoring functions and the margin loss. Remarkably, this is seamlessly transformed into a theoretically sound adversarial architecture, with no constrains on the hypothesis spaces of the classifier, achieving considerable improvements over state-of-the-art UDA methods.

While the MDD theory is sound for models with arbitrary hypothesis classes, it is unclear whether it is practical when applied to models with very large hypothesis spaces, such as those used for image segmentation.

Domain Adaptation for biomedical segmentation is currently not well understood theoretically. The theoretical understanding is particularly valuable in the medical domain, since it provides avenues for understanding capacity and limitations of the proposed methodology. This work seeks to investigate whether it is possible to apply MDD to the segmentation problem, by combining the U-Net \cite{ronneberger2015}, the architectural foundation for state-of-the-art medical segmentation models, with the MDD and propose a theoretically justified domain adaptation methodology for biomedical image segmentation. 

The paper contributes by proposing a new methodology, including a new training procedure with a novel early stopping scheme. The method is evaluated on the task of hippocampus segmentation of brain MRI. We find that the proposed methodology achieve an significant improvement on the base U-Net.

This work is to be considered a proof of concept, and provides an avenue to both further understanding, analyzing and applying Domain Adaptation to the medical domain. The theoretical justification of the proposed methodology, opens up completely new avenues of research, potentially providing fundamental contributions to our understanding of the capabilities and limitations of adversarial Domain Adaptation.

\begin{table*}[tb]
\setlength{\tabcolsep}{5pt}
\renewcommand{\arraystretch}{1.4}
\centering
\caption{Mean Dice scores with one standard deviation of 3D volumes on the Hippocampus datasets, with each experiment conducted $N = 6$ times.}
\label{tab:results}
\begin{tabular}{ll|cccc}
\toprule
\textbf{Source} & \textbf{Target} & \textbf{U-Net} & \textbf{MDD-UNet} \\ 
\midrule
\multirow{3}{*}{Hammers} & LPBA40 & $0.54\pm 0.01$ & $\mathbf{0.69 \pm 0.02}$ \\
 & Oasis & $0.65\pm 0.01$ & $\mathbf{0.71\pm 0.02}$ \\
 & HarP & $0.56\pm 0.02$ & $\mathbf{0.64\pm 0.01}$ \\ 
\multirow{3}{*}{HarP} & Hammers & $0.67\pm 0.003$ & $\mathbf{0.68\pm 0.01}$ \\
 & LPBA40 & $0.49\pm 0.01$ & $\mathbf{0.54\pm 0.03}$  \\
 & Oasis & $0.77\pm 0.01$ & $\mathbf{0.79\pm 0.01}$ \\ 
\multirow{3}{*}{Oasis} & Hammers & $0.64\pm 0.02$ & $\mathbf{0.69\pm 0.01}$ \\
 & LPBA40 & $0.28 \pm 0.1$ & $\mathbf{0.61\pm 0.05}$ \\
 & HarP & $0.6 \pm 0.06$ & $\mathbf{0.72\pm 0.03}$ \\
\multirow{3}{*}{LPBA40} & Oasis & $0.57\pm 0.01$ & $\mathbf{0.63\pm 0.01}$  \\
 & Hammers & $\mathbf{0.62\pm 0.01}$ & $0.60\pm 0.04$\\
 & HarP & $0.36\pm 0.03$ & $\mathbf{0.48\pm 0.09}$ \\
 \bottomrule
\end{tabular}
\end{table*}

\section{Method}

This section outlines the proposed method and it's theoretical foundation. We assume input space $\mathcal{X}$ and label space $\mathcal{Y}$, and two distributions $S$ and $T$ over $\mathcal{X} \times \mathcal{Y}$, denoted the \textit{source domain} and \textit{target domain} respectively. The model receives two samples: A labeled sample $\hat{S}$ from $S$ and an unlabeled sample $\hat{T}$ from $T_X$, the marginal distribution of $T$ over $\mathcal{X}$. The goal is to perform well on the target domain using only $\hat{S}$ and $\hat{T}$.

\subsection{Margin Disparity Discrepancy}

The theoretical foundation of the proposed method is the \textit{Margin Disparity Discrepancy} (MDD) from \citet{zhang2019}. Let $f : \mathcal{X} \to \mathbb{R}^{|\mathcal{Y}|}$ denote a classifier outputting a \textit{score} for each possible label and $\mathcal{F}$ a family of classifiers. Let $L^{(\rho)}_f : \mathbb{R}^{|\mathcal{Y}|} \times \mathcal{Y} \to \mathbb{R}$ denote the \textit{Margin Loss}, which measures how certain $f$ is at predicting $y \in \mathcal{Y}$ up to some maximal margin $\rho > 0$. This directly induces the \textit{Margin Disparity}, a measure of agreement between two classifiers $f$ and $f'$ measured by the margin loss. Let $h_f : x \mapsto \argmax_{y \in \mathcal{Y}} f(x)_{y}$. The Margin Disparity is then for some sample $\hat{D}$ of size $m$ from distribution $D$ over $\mathcal{X}$ given as
$$
\text{disp}^{(\rho)}_{\hat{D}}(f', f) := \frac{1}{m} \sum_{i=0}^{m} L^{(\rho)}_{f'}(f'(x_i), h_f(x_i)),
$$

a measure of how certain $f'$ is at predicting the same as $f$ averaged over the input image. \autoref{fig:disparity} shows how the Margin Disparity can be calculated from medical images by taking the average over all pixels. The Margin Disparity can be used to formulate a discrepancy metric between two distributions $D, D'$, over $\mathcal{X}$, called the Margin Disparity Discrepancy (MDD). The MDD is defined as
$$
\text{mdd}^{(\rho)}_{f, \mathcal{F}}(\hat{S}, \hat{T}) := \sup_{f' \in \mathcal{F}} \left( \text{disp}^{(\rho)}_{\hat{S}}(f', f) - \text{disp}^{(\rho)}_{\hat{T}}(f', f)\right).
$$
MDD measures discrepancy between two distributions as the \textit{maximal difference in expected margin loss} of any classifier $f' \in \mathcal{F}$ with respect to $f$. Importantly, Zhang et al. \cite{zhang2019} provides rigorous generalization bounds based on the MDD, showing that there is a trade-off between generalization error and the choice of margin. 

A central property of the MDD is that $\mathcal{F}$ can be a family of classifiers which is able to perform medical image segmentation. The MDD is optimizable by following Ganin et al. \cite{ganin} and introducing a feature transformation, $\psi$. Applying $\psi$ to the source and target domains, the overall minimization problem can be written as
\begin{align}
\label{eq:mdd_optimization_problem}
\min_{f, \psi} \text{err}_{\psi(\hat{S})}^{(\rho)}(f) + \text{mdd}_{f, \mathcal{F}}^{(\rho)}(\psi(\hat{S}), \psi(\hat{T})).
\end{align}
This is naturally a \textit{minimax game}, where the goal is to learn a representation, such that the final classification is based on features which are both discriminative and invariant to the change of domains.

\subsection{Network Architecture}

We combine the MDD with the U-Net. The U-Net is naturally split into \textit{blocks}, each consisting of one or more convolution operations and ReLU activation functions and combined using either max pool in the contracting path and up-convolution in the expanding path. We only consider models applied on 2D data, which can be obtained from 3D volumes by considering each slice independently. We apply MDD to the U-Net by splitting it into four parts:

\begin{enumerate}
\item $f_c$: The classifier. The top block of the expanding path, consisting of two convolution layers and ReLU activations and the final segmentation layer.
\item $f_a$: The adversary. An exact architectural copy of the classifier.
\item $f_d$: The decoder. All blocks on the expanding path except for the classifier and adversary.
\item $f_e$: The encoder. All blocks on the contracting path including the last block until the first up-convolution.
\end{enumerate}
We let $\psi = f_e \circ f_d$ and optimize \autoref{eq:mdd_optimization_problem} using an adversarial architecture. $\psi$ is trained through a Gradient Reversal Layer (GRL) (\autoref{sec:grl}). The architecture is given in \autoref{fig:architecture}.

\subsection{Gradient Reversal Layer}
\label{sec:grl}

The Gradient Reversal Layer (GRL) follows Ganin et al. \cite{ganin}. In the forward pass, the layer is simply the identity function. In the backwards pass the gradient is multiplied with a negative constant, $\eta$, effectively forcing $\psi$ to transform the input into domain invariant features by maximizing the MDD which minimise loss with respect to the parameters passed through the GRL.

\subsection{Loss}
Since the margin loss is prone to vanishing gradients, we follow \cite{zhang2019} and use the cross entropy loss to optimize \autoref{eq:mdd_optimization_problem}. Let $\sigma : \mathbb{R}^K \to (0, 1)^K$ denote the softmax. Let $N = 256^2 \cdot B$ where $B$ is the batch size. For source data $(x^s, y^s) \in \hat{S}$ the classifier, $f_c$, simply seeks to approximate $\text{err}_{\psi(S)}^{(\rho)}(f)$ using the standard cross entropy loss
\begin{align}
L^{c}(x^s, y^s) &:= -\frac{1}{N}\sum_{i = 1}^{N} \log\left[\sigma(f_c(\psi(x^s)))_{y^s}\right]_i.
\label{eq:loss_src_clf}
\end{align}
For target data $x^t \in \hat{T}$ the adversary seeks to approximate $\text{mdd}_{f, \mathcal{F}}^{(\rho)}(\psi(\hat{S}), \psi(\hat{T}))$ using
\begin{align}
&\begin{aligned}
L^{a'}(x^s) := -\frac{1}{N}\sum_{i = 1}^{N}\log\left[\sigma(f_a(\psi(x^s)))_{h_{f_c}(x^s)}\right]_i
\label{eq:loss_src_adv}
\end{aligned}\\
&\begin{aligned}
L^{a''}(x^t) :=
\frac{1}{N}\sum_{i = 1}^{N}\log\left[1-\sigma(f_c(\psi(x^t)))_{h_{f_c}(x^t)}\right]_i
\label{eq:loss_tgt_adv_sub}
\end{aligned}
\end{align}
where the modification of the cross entropy loss in \autoref{eq:loss_tgt_adv_sub} was introduced in \cite{goodfellow} to mitigate the adversarial burden of exploding or vanishing gradients. The total loss of $f_a$ is
\begin{align}
&L^a(x^s, x^t) := - L^{a''}_{\psi, f_a, f_c}(x^t) + \gamma\; L^{a'}_{\psi, f_a, f_c}(x^s).
\label{eq:loss_tgt_adv}
\end{align}
which is combined using a \textit{margin factor}, $\gamma = \exp \rho$. Note that the adversary is completely unsupervised, and instead depends on the classifier $f_c$. The margin factor $\gamma$ is treated as a hyperparameter, and is preferred relatively larger, however might lead to exploding gradients for large values. Note that \autoref{eq:loss_tgt_adv} is formulated as a minimization problem, and thus the total objective of the MDD becomes
\begin{align}
\min_{\psi, f_a, f_c} L^a_{\gamma, \psi, f_a, f_c}(x^s, x^t) + L^{c}_{f_c, \psi}(x^s, y^s),
\label{eq:loss_total}
\end{align}
which can be directly optimized using stochastic gradient descent.

\begin{table}[tb]
\centering
\caption{\textbf{Frozen layers.} Dice score on the target distribution by epoch on the validation split for different choices of freezing blocks. We define a block as the convolution layers with the same feature map size delimited by max-pool or up-convolution. We count the blocks from left to right, that is the order of the forward pass. }
\label{tab:frozen_layers}
\begin{tabular}{l|lll}
\toprule
\textbf{Frozen layers \textbackslash\ Epoch}      & \textbf{2} & \textbf{6} & \textbf{12} \\ 
\midrule
First encoder block                              & \textbf{0.6}        & \textbf{0.63}       & 0.68        \\
First 2 encoder blocks                         & \textbf{0.6}        & 0.62       & \textbf{0.69}        \\
First 3 encoder blocks                       & 0.6        & 0.61       & 0.63        \\
All of encoder                                   & 0.58       & 0.59       & 0.62        \\
Last 2 blocks of encoder                       & 0.48       & 0.28       & 0.38        \\
Last block of encoder + \\
first block of decoder                           & 0.45       & 0.36       & 0.12        \\
First 2 blocks in decoder                      & 0.36       & 0.01       & 0.0         \\
\bottomrule
\end{tabular}
\end{table}

\subsection{Pre-training and early stopping}
\label{sec:pretraining_early_stopping}
The MDD-UNet is trained by first training a standard U-Net on the source dataset. That is the model $f_c \circ f_d \circ f_e$, trained with the loss given in \autoref{eq:loss_src_clf}. MDD is then applied by copying the weights of $f_c$ into $f_a$ and training with the loss from the previous section, effectively treating the domain adaptation as a fine-tuning step.

Since MDD is applied to a model which has already learned to segment the source dataset \autoref{eq:loss_src_adv} is expected to be numerically quite close to zero. Interestingly, this metric is seemingly associated with performance degradation, and tends to increase rapidly right before performance degenerates. Thus, we introduce a new early stopping metric, which importantly can be performed without knowledge about the labels of $T$: Let $\xi > 0$, we then stop training immediately when
$$
    L_{\psi, f_a, f_c}^{a'}(x^s) > \xi,
$$
for some batch of source data $x^s$. In practice we used $\xi = 0.02$.

\section{Experimental setup}
We validate the effectiveness of the proposed methodology on the task of hippocampus segmentation.

\subsection{Data}

The core data used in this study are T1-weighted MRI volumes from \cite{ghazi2022}. Labels highlighted the hippocampus, separated into three class labels: left hippocampus, right hippocampus and background. The data consists of four datasets, which are used to represent distributional shift, by choosing different datasets as the source and target domains respectively. The datasets are:

\begin{enumerate}
\item \textbf{HarP}: 135 MRI scans from the ADNI study \cite{clifford2008} of normal, cognitively impaired, and demented subjects (65 female and 70 males) aged 60 to 90. Data was acquired using scanners from GE, Philips, and Siemens, with strengths of 1.5T or 3T.
\item \textbf{Hammers}: 30 MRI scans from \cite{faillenot2017} of healthy subjects (15 female and 15 male) aged 20 to 54. Data was acquired using a 1.5T GE scanner.
\item \textbf{Oasis}: 35 MRI scans from the MICCAI 2012 Multi-Atlas Labeling challenge \cite{landman2012, marcus2007} of healthy subjects (22 females and 13 males) aged between 18 and 90. Data was acquired using a 1.5T GE scanner.
\item \textbf{LPBA40}: 40 MRI scans from \cite{shattuck2008} of healthy subjects (22 females and 13 males) aged between 19 and 40. Data was acquired using a 1.5T GE scanner.
\end{enumerate}

\subsection{Preprocessing}
All volumes were skull stripped using the robust learning-based brain extraction system ROBEX \cite{iglesias2011}, bias field corrected and transformed to the RAS+ orientation. Moreover, the intensities of each volume were limited to the 99th percentile, standardized to have a zero mean and unit variance, and then scaled to the range $-1$ to $1$. Since the network only handles 2D input, volumes were sliced on the coronal dimension and padded to size $256 \times 256$.

\subsection{Model configurations}
The MDD-UNet is compared to the U-Net. U-Nets are first trained for $60$ epochs before MDD is applied. Models were trained using Adam \cite{kingma2014} with different learning rates for different parts of the MDD-UNets. When applying MDD, we freeze the first two layers of $f_e$, and apply early stopping with $\xi = 0.02$. When training MDD on top of U-Net trained with augmentations, the MDD did not use augmentations. U-Nets are trained with an initial learning rate of $10^{-3}$ which is subsequently halved every $10$ epochs. When applying MDD the learning rate of $f_a$ was $10^{-6}$, the learning rate of $f_c$ was $10^{-3}$, $f_e$ and $f_d$ was $10^{-3}$ multiplied by $\frac{2}{3}$ and $\frac{4}{9}$ respectively. The MDD-UNet used a margin factor of $\gamma = 0.08$, and a GRL constant of $\alpha = 1.4$.

\begin{figure}[tb]
\centering
\includegraphics[width=1\linewidth]{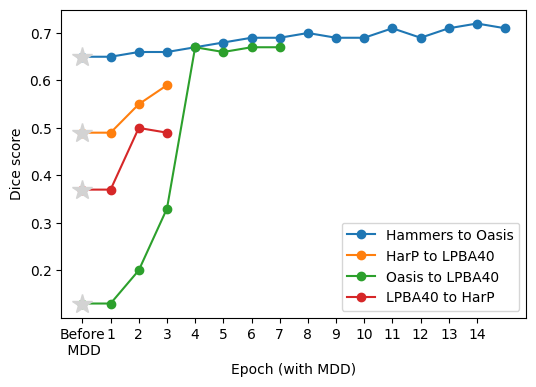}
\caption{\textbf{MDD Learning Curves}. Learning curves on the target domain for one run on different combination of source and target domains. Dice score is shown after each epoch of applying MDD and the length of each line denote when the training was stopped due to the early stopping mechanism, discussed in \autoref{sec:pretraining_early_stopping}. The learning curves are from one of the 6 runs performed in each combination of datasets. The plot clearly shows that the MDD is effective at mitigating performance loss due to domain shift, as the performance efficiently improves when the MDD is applied.}
\label{fig:mdd_finetune}
\end{figure}

\section{Results}
The results of our experiment are given in \autoref{tab:results}. The performance of MDD-UNet is a substantial improvement on the base U-Net, obtaining the best performance on 11 out of 12 combinations. 

\section{Discussion \& Limitations}

\textbf{Frozen layers}.
To analyze the impact of freezing layers, we conducted an experiment looking at the performance with different blocks frozen. We define a block as the convolution layers with the same feature map size delimited by max-pool or up-convolution. We count the blocks from left to right, that is the order of the forward pass. \autoref{tab:frozen_layers} denotes the dice score on the target distribution by epoch in the training progress on the validation split. The performance of the U-Net before adding MDD was 0.54 Dice on the target set. Freezing the first two blocks of encoder outperforms all other configurations, in particular any configuration where blocks in the decoder is frozen.

The frozen layers of the MDD-UNet indicate that the low-level features of the model are more domain invariant in the U-Net than the high level features. Further, since the hypothesis space of the max-player is incredibly large, it can be difficult to find the desired equilibrium between the adversaries. These results showcase how a combination of frozen layers in the beginning of $\psi$ and pre-training, achieves a stable training which allows the MDD to be applied with the early-stopping mechanism.
\\
\\
\noindent\textbf{Effectiveness of the MDD}. 
When the MDD is applied, the performance of the network on the target domain is efficiently improved. \autoref{fig:mdd_finetune} shows the learning curves as measured by Dice performance on the target domain by number of epochs of MDD application. When the MDD is applied, the target performance improves dramatically in only a few epochs. The early stopping mechanism reliably stops training exactly when the target performance is best or close to.
\\
\\
\noindent\textbf{Limitations.} 
This work does not claim to establish the MDD-UNet as a state-of-the art domain adaptation methodology, and future work should investigate the interplay with augmentations and other methodological improvements known to improve performance in the presence of domain shift \cite{chen2019synergistic}. Further, in this work we focused on demonstrating the effectiveness of the MDD on models working on 2D data. It is left for future work to investigate how the methodology behaves on 3D data, which is common in the medical domain, and modern adaptations of the U-Net \cite{lee2023d, isensee2021}.

\section{Conclusion}

In this paper we proposed a domain adaptation methodology with theoretical guarantees based on the U-Net and the MDD. We show that the MDD-UNet outperforms the regular U-Net for segmenting hippocampus data. This work opens the door for further studying the applications of the proposed methodology and importantly the MDD discrepancy metric to the biomedical domain. Further, this work opens the door for analysing biomedical Domain Adaptation theoretically, a completely new avenue of research in the biomedical field.

\section*{Acknowledgments}

This work was supported by Danish Data Science Academy, which is funded by the Novo Nordisk Foundation (NNF21SA0069429) and VILLUM FONDEN (40516), and by the Pioneer Centre for AI (DNRF grant number P1).

\printbibliography

\end{document}